\documentclass{article}
\usepackage{ijcai25}

\usepackage{times}
\usepackage{soul}
\usepackage{url}
\usepackage[hidelinks]{hyperref}
\usepackage[utf8]{inputenc}
\usepackage[small]{caption}
\usepackage{graphicx}
\usepackage{amsmath}
\usepackage{amsthm}
\usepackage{booktabs}
\usepackage{algorithm}
\usepackage{algorithmic}
\usepackage[switch]{lineno}
\usepackage{graphicx}   % For \scalebox
\usepackage{array}      % For custom column types like m{}
\usepackage{multirow}   % For \multirow
\usepackage{booktabs}   % For \toprule, \midrule, \bottomrule
\usepackage{colortbl}   % For \rowcolor
\usepackage{xcolor}     % For defining colors

\usepackage{pifont} % 使用该包提供的符号
\usepackage{longtable}
\usepackage{xspace}
\usepackage{footnote}

\title{QiMeng-TensorOp: Automatically Generating High-Performance Tensor Operators with Hardware Primitives}

\newcommand{\name}{QiMeng-TensorOp\xspace}
\author{
Xuzhi Zhang$^1$ \and 
Shaohui Peng$^1$ \and 
Qirui Zhou$^2$ \and 
Yuanbo Wen$^2$ \and 
Qi Guo$^2$ \and 
Ruizhi Chen$^1$ \and 
Xinguo Zhu$^1$ \and 
Weiqiang Xiong$^1$ \and 
Haixin Chen$^2$ \and 
Congying Ma$^3$ \and 
Ke Gao$^1$ \and 
Chen Zhao$^1$ \and 
Yanjun Wu$^1$ \And 
Yunji Chen$^2$ \and 
Ling Li$^{1}$\footnote{Corresponding author: \textit{Ling Li}}\\
\affiliations
$^1$Institute of Software Chinese Academy of Sciences\\
$^2$Institute of Computing Technology, Chinese Academy of Sciences\\
$^3$Peking University\\
\emails
\{zhangxuzhi2023, pengshaohui, ruizhi, gaoke, zhaochen, yanjun, liling\}@iscas.ac.cn,
\{zhuxinguo23, xiongweiqiang20, chenhaixin24\}@mails.ucas.ac.cn, 
\{zhouqirui22s, guoqi, cyj\}@ict.ac.cn, 
wenyb@mail.ustc.edu.cn, 
2100013182@stu.pku.edu.cn
}

\begin{document}

\maketitle

\begin{abstract}  

% Computation-intensive tges large languaconstitute over 90\% of the computations in Large Language Models (LLMs) and Deep Neural Networks (DNNs), thus ge models to traperformance.
    % nd search modules, LLM-AGheterogeneity and complexity developmentestensor operator implementationhardware adaptabhardwaresiciently and cost-effectively.
% s code genration taks generating hardward-primitive levelr, but still face chanllenges comhensive undestanding hardware charticas and finegrained optimization
    % LLM  frameworkactivate LLMs to automtic understanding hardware information to genrate hardward primitive-level tensor operator, and fine-grained tuning parameter to obtain optimal performance in target hardware.
    % Verified in a wide range
    Computation-intensive tensor operators constitute over 90\% of the computations in Large Language Models (LLMs) and Deep Neural Networks.
    % It is of crucial importance to automatically and efficiently generate high-performance tensor operators with hardware primitives, so that fully utilizing diverse and evolving hardware architectures (e.g. RISC-V, ARM and GPUs), which still chanllenging for exsiting handcraft optimized libraries due to high development cost and limited portability.
    % However, implementing tensor operators across various hardware platforms remains challenging due to high development cost and limited portability of existing handcrafted libraries.
    Automatically and efficiently generating high-performance tensor operators with hardware primitives is crucial for diverse and ever-evolving hardware architectures like RISC-V, ARM, and GPUs, as manually optimized implementation takes at least months and lacks portability.
%    as existing hand-optimized libraries face challenges due to high development costs and limited portability.
    %LLMs demonstrate outstanding performance in generating 
    LLMs excel at generating high-level language codes, but they struggle to fully comprehend hardware characteristics and produce high-performance tensor operators.\\
    %As LLMs become increasingly capable of code generation tasks, they offer potential for generating hardware-primitive-level tensor operators.
    %Nonetheless, they still face challenges in comprehensively understanding hardware characteristics and performing fine-grained optimization.
    %To address these issues, we introduce LLM-AGTOP, an LLM agent framework that enables LLMs to automatically understand hardware information, generate hardware-primitive-level tensor operators, and fine-tune parameters to achieve optimal performance on target hardware.
    %To address these issues, we introduce an LLMs framework (\name), which empowers LLMs to automatically utilize hardware characteristics to generate hardware-primitive-level tensor operators, and tuning parameters to achieve optimal performance on various target hardware platforms.
    We introduce a tensor-operator auto-generation framework with a one-line user prompt (\name), which enables LLMs to automatically exploit hardware characteristics to generate tensor operators with hardware primitives, and tune parameters for optimal performance across diverse hardware.
    Experimental results on various hardware platforms, SOTA LLMs, and typical tensor operators demonstrate that \name effectively unleashes the computing capability of various hardware platforms, and automatically generates tensor operators of superior performance. 
    Compared with vanilla LLMs, \name achieves up to $1291 \times$ performance improvement. Even compared with human experts, \name could reach $251 \%$ of OpenBLAS on RISC-V CPUs, and $124 \%$ of cuBLAS on NVIDIA GPUs. Additionally, \name also significantly reduces development costs by $200 \times$ compared with human experts.
     %这里的数据到底多少？
    %$100\times$ improvement over vanilla LLMs and at most $2\times$ improvement over handcraft libraries and auto-compilers. Additionally, AGTOP also significantly reduces development costs (12$\times$ faster) compared with auto-compilers and handcreaft libraies.onstitute over 90\% of the computations in
\end{abstract}

\section{Introduction}
\begin{figure}[t]
    \centering
    \includegraphics[width=0.85\linewidth]{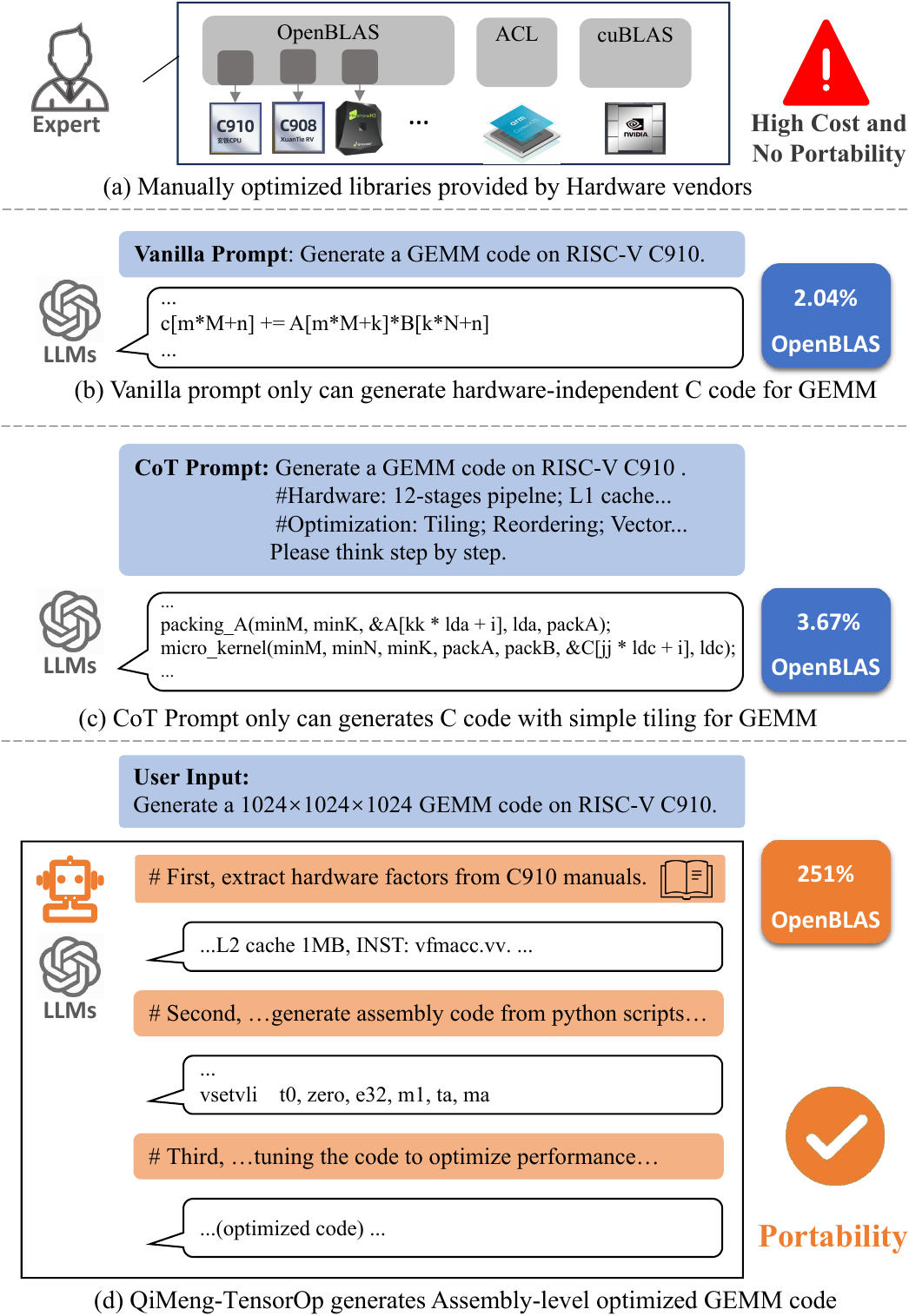}
    \caption{Comparison of tensor operator optimization paradigms}
    \label{fig1:Intro}
\end{figure} Tensor operators, like General Matrix Multiplication (GEMM) and Convolution (Conv) \cite{chenLearn}, are critical in various mathematical and computational fields, especially in deep learning, as they constitute over 90\% of the computations in LLMs and Deep Neural Networks(DNNs) \cite{gemm_occupy1,gemm_occupy2}.
% Deep Neural Networks (DNNs)~
The significance and unique computational requirements of tensor operators have driven heterogeneity and complexity in hardware design, such as Tensor Core in NVIDIA GPU \cite{tensorcore}, RISC-V vector extension (RVV), and deep learning accelerators like Google TPU \cite{tpu} and Cambricon \cite{cambricon}.
% hardware-primitive-level implementation and optimization, like assembly instructions for CPU and hardware intrinsic for CUDA Tensor core,  which present new challenges for deploying and optimizing tensor operators effectively.

%Implementation and optimization at the hardware primitive level, like assembly instructions for CPU and hardware intrinsic for CUDA Tensor core, are crucial for the portability of tensor operators and for fully leveraging hardware performance.
Implementing tensor operators with hardware primitives (such as assembly instructions for CPU and hardware intrinsic for CUDA Tensor core) is the only way to fully maximize hardware performance~\cite{xiao2021hasco,liu2022high,igual2023automatic,alaejos2024automatic,castello2024tackling,wu2024autogemm}.
Hardware primitives provide programmers with precise control over hardware resources (including computing units, registers, memory, etc.), thereby enabling exceptional performance~\cite{tlm}.
For instance, 
%This is because such primitive can more thoroughly and finely control and leverage the underlying characteristics of the hardware, 
an assembly implementation of GEMM can yield over 62,000$\times$ performance improvement compared to the vanilla Python implementation \cite{hennessy2019new}.
% Therefore, fully optimizing tensor operators at the hardware-primitive-level for different hardware platforms is essential to maximize their performance.
%However, due to their high correlation with hardware characteristics, hardware primitives have a high level of abstraction and complexity, making them very difficult for programmers to use, let alone to generate automatically.

%Developing with hardware primitives is highly challenging, as it requires a profound understanding of hardware architectures. The development process is intricate, often resulting in low efficiency and a high error rate, which demands significant proficiency from programmers. 

However, designing high-performance tensor operators with hardware primitives is challenging, as it requires a deep comprehension of hardware architectures. 
Besides, it is intricate and often results in low efficiency and high error rates. 
Existing tensor operators are mainly implemented with two paradigms, manually optimized libraries and auto-compilers. 
% Currently, tensor operator is primarily implemented through hand-optimized libraries and auto-compilers. %Manually optimized verse hardwlikelatforms. 
%^To tackle this, researchers are increasingly turprovided by hardware vendors are manually implemented by experts using hardware primitives and specific optimizations for different platforms. Manually optimized libraries not only require at least months for experts to optimize tensor operators, that is extremely time-consuming, but also lack portability across platforms.
Manually optimized libraries provided by hardware vendors, such as MKL \cite{MKL} and ACL for CPUs, cuBLAS and cuDNN for GPUs, are developed by human experts using hardware primitives and specific optimizations for various platforms. This process is time-consuming, often taking months to optimize tensor operators, and these libraries also lack portability across different platforms.
Auto-compilers, such as Halide \cite{halide}, TVM \cite{tvm}, and Ansor \cite{ansor,metaschedule,bi2023heron,tlp,tlm}, 
% also need experts to predefine hardware-specific rules and backends to search and deploy tensor operator implementations.
explore a vast program space to generate efficient tensor operators. However, they still require human experts to manually define hardware-specific rules and backend implementations to optimize and deploy tensor operations.
Despite alleviating certain manual labor, they necessitate significant expertise and involve high development costs, 
%demands a high degree of expertise and entails substantial development costs, 
as exemplified by the arduous challenge of deploying compilers such as TVM on RISC-V CPUs \cite{tvmriscv}.Consequently, existing paradigms struggle with development cost and portability, making them insufficient to keep up with rapid hardware advancements.

LLMs (GPT-4o~\cite{gpt4o}, DeepSeek-V3~\cite{DeepSeekV3}) have achieved remarkable progress in code generations, making  natural language to code (NL2Code) one of the most popular paradigms~\cite{NL2Code2,NL2Code3,NL2Code1}
% gpt4o} code generations,%
% and 
   % ,NL2Code1
However, existing LLM-based code generation researches \cite{ReACC,StarCoder,CodeLlama,Textbook} mainly focus on hardware-independent high-level language code generation, and are 
% LLM-based code generation pays focus on high-level language generation \cite{StarCoder,CodeLlama,Textbook,ReACC}, and is almost 
incapable of hardware-primitive-level code generation. LLMs yet struggle to fully comprehend hardware characteristics and correctly manipulate hardware resources to implement and optimize assembly code, let alone high-performance tensor operator generation with simple prompts. 

%A new paradigm that automatically generates assembly-level tensor operators leveraging LLMs, which can quickly adapt to for different hardware platforms to maximize hardware performance and address portability challenges.%However, current methods for code generation pays focus on high-level programming language to complete functionality and are almost incapable of handling the more abstract and complex 
%hardware-primitive-level generation. \cite{StarCoder,CodeLlama,Textbook,ReACC}.
% Additionally, tensor operators demand that LLMs have comprehensive understanding and application of the relation between hardware characteristics and optimization techniques.
%Vanilla prompts or 
% Chain of Thought (CoT) also leads to very poor performance, rendering impractical for user to achieve high-performance tensor operator implementations with just give LLMs a simple description (see Figure 1\ref{fig1:Intro}).
% Chain of Thought (CoT) also results in poor performance, making it impractical for users to achieve high-performance tensor operator implementations with hardware primitives by simply providing LLMs with a basic description (see Figure \ref{fig1:Intro}).majorautomatically generating -s.
% The first challenge is to make LLMs understand hardware architectures and use hardware primitives to accurately implement tensor operators.
In summary, we observe two major challenges for LLMs in automatically generating hardware-primitive-level tensor operators.
The first challenge is enabling LLMs to comprehend hardware architectures and accurately utilize hardware primitives to implement tensor operators.
%The second challenge is to optimize the performance of generated tensor operators with hardware intrinsic optimization techniques.
The second challenge is optimizing the performance of the generated tensor operators through hardware intrinsic optimization techniques.

To address the challenges, we propose the \name framework that automatically generates high-performance tensor operators with hardware primitives on various platforms. \name only needs a one-sentence prompt from user, which describes the target tensor operator and hardware, as shown in Figure \ref{fig1:Intro}. \name uses general hints to trigger LLMs to comprehend hardware optimization and auto-extract target hardware factors. Then it leverages LLMs and extracted factors to generate sketch and kernel codes with hardware primitives for target operators. Finally, it exploits LLMs' in-context learning via MCTS to uncover the optimization opportunities for generated codes.
% \name consists of three parts: Hardware Architecture Understanding, which uses general hints to trigger LLMs' hardware optimization understanding and auto-extracts target hardware factors; Tensor Operator Generation, leveraging LLMs and extracts factors to create sketch and kernels with hardware primitives; Auto-Tuning, exploiting LLMs' in-context learning via MCTS to uncovering subtle optimization opportunities.
In short, \name effectively harnesses LLMs' knowledge and capabilities to comprehend and apply optimization techniques on various hardware platforms, enabling efficient auto-generation and tuning of tensor operators with hardware primitives.

 To our knowledge, we are \textit{the first to automatically generate high-performance tensor operators with hardware primitives by exploiting LLMs}. This paper makes the following contributions: 
\begin{itemize}
    %\Tse a novel paradigm that automatically generates assembly-level t% ensor operators for different hardware to achieve both performance and portability
    % We also identify LLM agents as a potential solution and outline the key challenges it entails.
    % \item We introduce a two-stage LLM agent framework that effectively resolves the semantic alignment issue of LLM generating assembly tensor operators and addresses the challenge of efficient fine-grained optimization.
    % \item Verified on a wide range of hardware (ARM and RISC-V CPUs, NVIDIA GPUs) for tensor operators (GEMM and Conv), \name demonstrated more than $100\times$ improvement over vanilla LLMs and at most $2\times$ improvement over handcraft libraries and auto-compilers. Additionally, AGTOP also significantly outperformed the automatic compiler TVM and other search algorithms in optimization efficiency.
    % \item Finally, the proposed \name framework significantly reduces development costs: it is over 12$\times$ faster than an auto-compiler like TVM (20 minutes versus 4 hours) and 120$\times$ faster than a senior coder (20 minutes versus 5 workdays).
    \item We propose a framework for automatically generating tensor operators at the hardware primitive level across various platforms, requiring only a single sentence from users to describe the target operator and hardware.%, allowing LLMs to efficiently produce and tuning high-performance tensor operators across various platforms.
    \item We develop general hardware intrinsic optimization hints and workflow to help LLMs comprehend hardware and optimization techniques, allowing to automatically extract information from manuals to generate tensor operators with hardware primitives.
    \item We design an LLM-assisted MCTS algorithm that effectively enhances the efficiency and performance of tuning primitive-level tensor operators on specific hardware.
    \item Extensive evaluations across diverse hardware platforms and tensor operators (GEMM and Conv) of various dimensions demonstrate \name significant performance promotion (up to 1291$\times$ of vanilla prompt and up to $251\%$ than manually optimized libraries) and development cost reduction (200$\times$ than senior coder).
\end{itemize}
\section{Preliminary}
\subsection{Tensor Operator}
%Computation-intensive Tensor operators in deep learning, 
GEMM and Conv, the most important tensor operators, are computation-intensive. 

\textbf{GEMM} 
% is a fundamental operation in linear algebra and numerical computing that 
refers to the multiplication of two dense matrices, $A \in R^{m\times k}$ and $B \in R^{k\times n}$, as
% is an essential operation in linear algebra and constitutes the majority of computations in deep learning applications, especially in LLM inference. The formula for GEMM is as follows:
$(AB)_{ij}=\sum_{q=1}^kA_{iq}\cdot B_{qj}$.%$, A \in R^{m\times k}$ and $B \in R^{k\times n}$

\textbf{Conv} slides a filter ($K$) on an input ($X$), and calculates element-wise dot product, $Y(i, j) = \sum_{m=0}^{M-1} \sum_{n=0}^{N-1} X(i+m, j+n) \cdot K(m, n)$. %between the filter and the input
% is a fundamental operation in signal processing, image processing, and neural networks, which 
% involves the application of a filter ($K \in R^{M\times N}$) to an input to produce an output, as:
% is a core operation in the field of computer vision, widely used in tasks such as image processing, feature extraction, and object recognition.  The formula for 2D image convolution can be expressed as:
% \begin{equation}
% Y(i, j) = \sum_{m=0}^{M-1} \sum_{n=0}^{N-1} X(i+m, j+n) \cdot K(m, n)
% \end{equation}
Conv is commonly implemented by converting the input and filter tensors into 2D matrices with Image-to-Column and then calling GEMM in BLAS libraries.
% Where: \(X(i+m, j+n)\) is the pixel value at position \((i+m, j+n)\) in the input image; \(K(m, n)\) is the value at position \((m, n)\) in the convolution kernel; \(M\) and \(N\) are the height and width of the kernel, respectively.
% We  GEMM (General Matrix Multiplication) to implement convolution (Conv) is a common optimization technique in deep learning frameworks and libraries.
% A common implementation of convolution is to transform the convolution operation into a matrix multiplication operation. Specifically, this is achieved by rearranging the convolution operation into a matrix operation using the \texttt{Im2col} method, thereby leveraging highly optimized GEMM  to accelerate convolution computation. 
% We use the Image-to-Column (Img2Col) to convert Conv input and filter tensors into 2D matrices. This widely used strategy can implement Conv with GEMM, leveraging advanced optimization techniques and hardware characteristics.
% This widely used strategy can implement Conv with GEMM, leveraging advanced optimization techniques and hardware characteristics.

\begin{figure*}[t]
    \centering
    \includegraphics[width=1\linewidth]{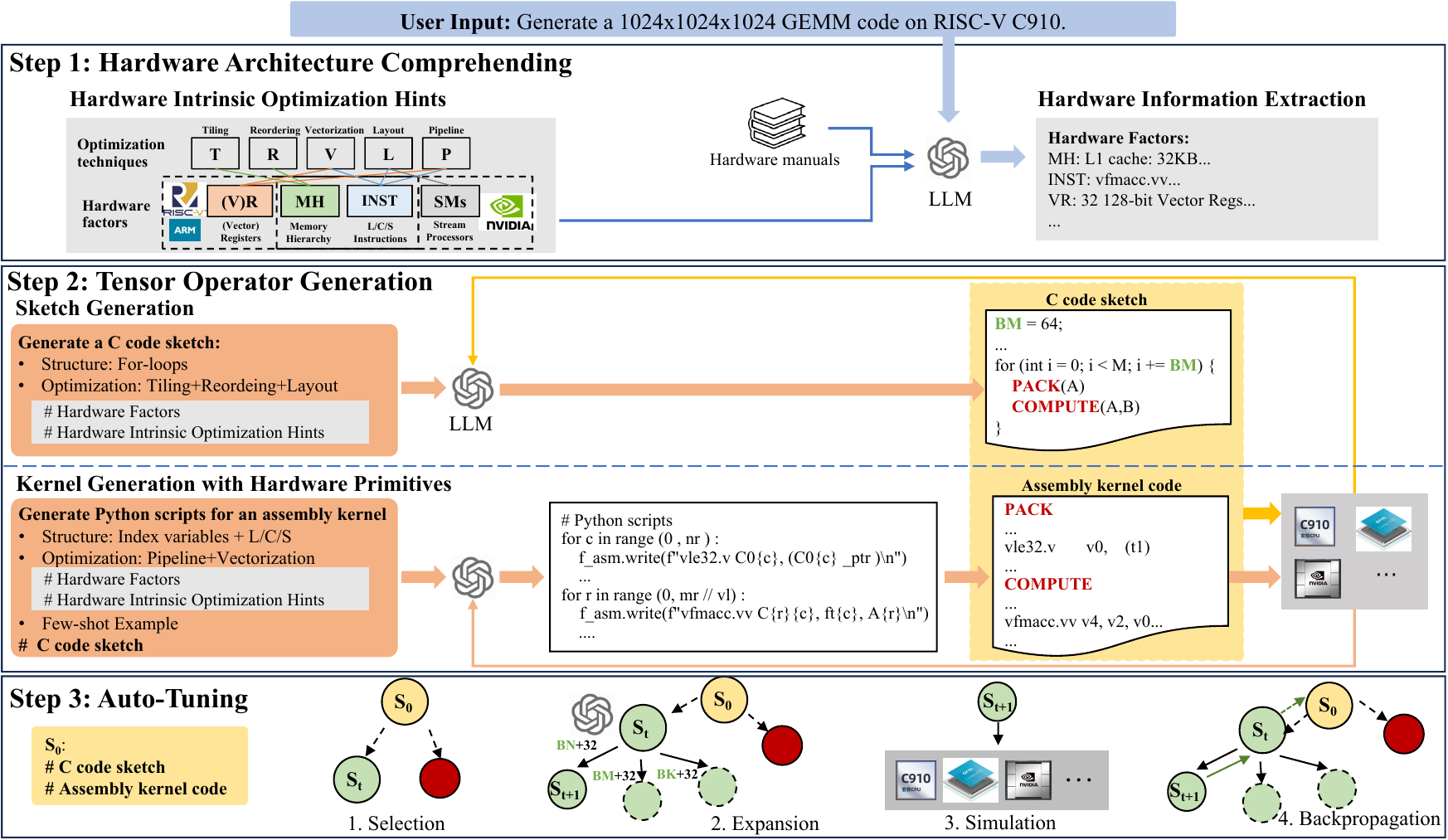}
    \vspace{-5 pt}
    \caption{\textbf{\name overview.} The proposed framework takes a user's one-sentence description as input and generates high-performance tensor operators using hardware primitives via three automatic steps. Step 1 activates LLMs’ comprehension of tensor operator optimization and extracts target hardware factors for subsequent generation. In Step 2, sketches and kernels are generated to form tensor operators. Sketch generation uses optimizations such as tiling, and hardware-primitive-level kernels like \textit{PACK} and \textit{COMPUTE} employ instruction-level optimization. In Step 3, Auto-Tuning uncovers subtle optimization opportunities and further enhances performance. (A detailed showcase please refer to Appendix C.)}
    \label{fig:overview}
    \vspace{-5 pt}
\end{figure*}

\subsection{Optimization Techniques}
%There are generally five kinds for tensor operator implementation: 
The optimization techniques for tensor operators can be essentially classified into several fundamental operations across hardware architectures.
% \begin{itemize}
% \item Tiling: decomposing a large matrix into smaller matrix computation subproblems to maximize the use of cache for element access.
% \item Reordering: reordering the for-loops to optimize memory access efficiency.
% \item Vectorization: using vector instructions to parallelize the loading and computation of multiple elements.
% \item Layout: using vector instructions to parallelize the loading and computation of multiple elements.
% \item Pipeline: improving the utilization of functional units by rearranging the computation and memory access instructions.
% \end{itemize}
% \begin{itemize}

\textbf{Tiling (T)} decomposes a  matrix into smaller blocks to improve memory access~\cite{faingnaert2021flexible}.
%into smaller matrix computation subproblems to maximize the use of cache for element access.

\textbf{Reordering (R)} exchanges a for-loops to boost memory access efficiency~\cite{anam2013precision}.
% reordering the for-loops to optimize memory access efficiency.

\textbf{Vectorization (V)} packs matrix data to use vector instructions for computation and memory access~\cite{katel2021high}.
% changes the arrangement of the matrix to match specific hardware memory access characteristics.

\textbf{Layout (L)} rearranges matrix data to better fit the hardware's memory access patterns~\cite{kurzak2012autotuning}. 
% using vector instructions to parallelize the loading and computation of multiple elements.

\textbf{Pipeline (P)} overlaps the computation and memory access to minimize the memory access latency~\cite{tan2011fast}.
% enhances performance by overlapping various stages of computation and data movement, allowing for seamless processing and minimizing idle time between operations~\cite{tan2011fast}.

The five optimization techniques vary in implementation challenge~\cite{feng2023tensorir}.
% Vectorization, Layout and Pipelining require hardware primitives for specialized data movement and computation.
% In contrast, Tiling and Reordering involve designing high-level task and data allocation strategies to match with hardware characteristics.
Vectorization, Layout, and Pipeline need hardware primitives for specialized data movement and computation, while Tiling and Reordering involve designing high-level task and data allocation strategies to fit hardware characteristics.

\subsection{Hardware Factors}\label{sec:hw-factors}
% To fully leverage hardware performance, we 
We summarize four key hardware factors essentially impacting the implementation and optimization of tensor operators.
% that has essential impact on the implementation and optimization of tensor operators.

% The 5 optimization techniques has different generation difficulty.
% Vectorization, Pipeline and Layout needs instruction-level hardware primitive to implement special data movement and computation, while the last Tiling and Reordering need design high-level task and data allocation to match hardware characteristics. 
%the implemetation Tensor operators must fully leverage hardware characteristics to achieve optimal performance.
%To validate the generalizability of our approach, we conducted extensive experiments on various platforms, including CPUs with ARM and RISC-V, as well as GPUs with CUDA cores and Tensor cores.

\textbf{Memory Hierarchy (MH)} refers to how hardware organizes and manages different levels of memory, such as L1/L2/L3 cache structures of CPUs~\cite{wu2021storage} and the global/shared memory of NVIDIA GPUs~\cite{dally2021evolution}. Memory Hierarchy is crucial for efficient memory access and optimizations like Tiling and Reordering.
% CPU的存储层次主要包含内外存，多级cache和寄存器，其中多级cache的结构对于优化张量算子的访存效率非常重要。
% GPU的存储层次主要包含全局/共享存储，多级cache，其中全局/共享存储可由用户控制，对于张量算子的数据复用，流水线实现等关键优化必不可少。
% The memory hierarchy of CPUs primarily includes memory, multi-level caches, and registers, where the structure of multi-level caches is crucial for optimizing the memory access efficiency of tensor operators.
% The memory hierarchy of GPUs primarily includes global/shared memory and multi-level caches, where the global/shared memory can be controlled by the user and is essential for key optimizations such as data reuse and pipelining in tensor operators.

\textbf{Instructions (INST)}
% 不同硬件的指令集类型和用法决定了对于张量算子硬件原语级别实现的高效性，以及向量化、流水线优化等优化技术的应用。例如RISC-V架构的RVV指令，ARM架构的NEON指令以及NVIDIA GPU的CUDA Templates (CuTe).
are the basic operation units of computation and data movement, 
such as the RVV instructions in the RISC-V architecture, NEON instructions in the ARM architecture, and CUDA Templates (CuTe) of NVIDIA GPU Tensor Cores. Instructions determine the hardware primitives for tensor operator implementation ~\cite{xiao2021hasco}, and optimization techniques like Vectorization.
%, determine the implementation details of tensor operators at the hardware primitive level~\cite{xiao2021hasco}, as well as the application of optimization techniques like vectorization.

\textbf{Vector/Scalar Registers of CPUs ((V)R)} refers to the number and width of tensor/scalar registers. They are crucial for data movement and computation efficiency in tensor operators, affecting the generation of vector instructions and the granularity of Pipeline optimizations.
% CPU中的张量/标量寄存器的个数，宽度对于张量算子高效地利用搬运数据和计算十分重要，影响了张量运算指令的生成以及流水线优化的粒度等。

\textbf{Streaming Processor Information of GPUs (SMs)}
% 包括SP个数，SP中的cuda core和tensor core个数等，决定了生成cuda计算kernel时的grid和block计算任务分配维度和数据tiling方式。
includes the number of SMs, and the number of CUDA Cores and Tensor Cores within each SM~\cite{choquette2021nvidia}. It determines the grid and block dimension task allocation and data tiling when generating CUDA kernels.

\section{Method}
% 此部分讲清楚分成Semantic-level、Non-Semantic-level这两个level的优势，作为核心创新点之一。
% In this section, we introduce \name framework, which enables LLMs to understand hardware architectures, automatically generate high-performance tensor operators with hardware primitives, and optimize the generated code using MCTS. As shown in Figure \ref{fig:overview}, \name only needs one-line prompt from the user, and consists of three key components
% %automatically extracts target hardware information, generates tensor operators with hardware primitives, and optimizes them using MCTS leveraging LLMs.
% to be discussed in the subsections. 
% %We will introduce the three main process, Hardware Factor Extraction, Tensor Operator Generation and Auto-tuning module, in the subsequent subsections.
% %(More details refers to Appendix))

In this section, we introduce the \name framework. It enables LLMs to comprehend hardware architectures, auto-generate high-performance tensor operators with hardware primitives, and optimize the generated code via MCTS. As depicted in Figure \ref{fig:overview}, \name requires just a one-line user prompt and comprises three key components (a detaled).

% \vspace{-8 pt}
%\subsection{Hardware Factor Extraction}
\subsection{Hardware Architecture Comprehending}
The Hardware Architecture Comprehending consists of two parts. 
The Hardware Intrinsic Optimization Hints activate LLMs' comprehension of hardware and optimization, guiding subsequent generation, while the Hardware Factor Extraction ensures the exploitation of target hardware characteristics to further implement cross-platform optimization automation.
% the core hardware factors for tensor operator so that it can retrieve target hardware information from manuals and guild the following automatic tensor operator generation.

% \subsubsection{Hardware Intrinsic Optimization Hints}
\textbf{Hardware Intrinsic Optimization Hints.}
These hints serve as the background knowledge for LLMs to automatically conduct hardware factor extraction and tensor operator generation.
As in Figure \ref{fig:overview}, we summarize the description of five general tensor operator optimization primitives and their relationship with four key hardware factors (as described in Section \ref{sec:hw-factors}).
For example, the cache hierarchy of CPUs determines the tiling size of the matrix dimensions to ensure the locality of input data, thereby improving memory access efficiency.
For GPUs, the information of SMs determines the size of grid-dimension and block-dimension for the matrix computation to be allocated, thus enhancing the computational efficiency of the CUDA kernel.
The optimization hints are described with nature language, and thus developers can add knowledge to enable more hardware characteristics and optimization techniques conveniently. 

% \subsubsection{Hardware Factor Extraction}
\textbf{Hardware Factor Extraction.}
Hardware Factor Extraction takes users' prompt of tensor operator type and the target hardware name as input, and then efficiently retrieve information about the hardware factors of target hardware from the pre-collected set of manuals or those provided by the user (optional).
% This process aims to prompt LLMs to fulfill user requirements regarding the target tensor operator and target hardware.
% It is based on user input and our pre-collected hardware manuals.
% The user merely needs to input the tensor operator type (GEMM or Conv), shape, and the target hardware name.
% Subsequently, the LLMs can efficiently retrieve information about the hardware factors from the pre-collected set of manuals or those provided by the user (optional).
As depicted in Figure \ref{fig:overview}, hardware factors of RISC-V C910 include the cache levels and size, the type and usage of vector computation instructions, and so on.
Based on retrieved hardware factors, the subsequent process can generate targeted tensor operator implementation.

\subsection{Tensor Operator Generation}
%A Code Generation Agent takes a natural language description of the optimization strategy for operators on a specific platform as input. Through an iterative optimization process, it gradually generates a scheduling template and an assembly-level kernel.
% At the generation module, three processes are proposed to enable LLMs to understand and extract hardware characteristics to generate high-performance hardware-primitive-level tensor operator on the target platform.
%These steps allow us to obtain an initial solution on the target platform.
This component generates hardware-primitive-level tensor operators with proper optimization techniques.
Considering the different implementation challenges of optimization techniques, Vectorization, Pipeline require hardware primitives for optimized data movement and computation, while Tiling, Reordering and Layout involve task and data allocation strategies to match target hardware.
Thus, Tensor Operator Generation consists of Sketch Generation and Kernel Generation with Hardware Primitives as shown in Figure \ref{fig:overview}.

\textbf{Sketch Generation.}
Sketch Generation leverages LLMs to generate the main function (C for CPU, CUDA C++ for GPU) of tensor operators with task and data allocation optimization for better memory access efficiency.
It reserves indivisible computation or data movement as kernel calls  (e.g., \textit{PACK} and \textit{COMPUTE} for CPUs) to subsequent hardware-primitive-level generation.
The prompt delineates basic sketch structures and optimization techniques, including Tiling, Reordering, and Layout.
For CPUs, the basic structure is a simple three-level for-loop.
Specifically for the C910 CPU, LLMs utilize the prompt in conjunction with hardware factors and optimization hints to generate the main function featuring multi-level for-loops.
This implementation enables scheduling optimization tailored to the C910's characteristics while preserving PACK and COMPUTE kernels for subsequent hardware-primitive-level optimization.
For GPUs, LLMs produce CUDA C++ code that defines grid and block dimensions to efficiently allocate computational resources and execute CUDA kernels.
% To enhance parameters in sketch initialization for the next Step Auto-Tuning, the sketch code is compiled with kernels to test performance on the target hardware and adjust parameters using LLMs in a few cycles.

% \subsubsection{Kernel Generation with Hardware Primitives}
\textbf{Kernel Generation with Hardware Primitives.}
% As depicted in Figure \ref{fig:overview}, given that 
LLMs perform well in generating Python code, but they face challenges in generate functionally correct codes with hardware primitives (e.g., CPU assembly instructions or GPU PTX). Thus, we prompt LLMs to generate Python scripts as bridge for kernels.
The prompt comprises sketch code, the description of kernel structure and optimizations, few-shot examples, along with optimization-related hints and target hardware information acquired in Step 1.
For instance, when generating a computation kernel with Pipeline optimization for CPUs, the structure section delineates vector register definitions and the layout of Load, Compute, and Store instruction blocks.
Along with the target C910 CPU information (the type and usage of assembly INST, number and width of (V)R, etc.) in prompt, LLMs write a Python script to print the assembly-level kernel implementation.
Conversely, for GPUs, LLMs write CuTe code to implement CUDA kernels with PTX-level optimization, aiming to fully exploit the Tensor Core's peak performance.
To improve sketch parameter initialization for the subsequent Auto-Tuning step and the correctness of assembly kernels, they are jointly compiled and tested on the target hardware to obtain feedback for refining sketches and Python scripts of kernel generation.

% As shown in Figure\ref{fig:overview}, since LLMs struggle with precisely usage of hardware primitives (like assembly instructions of CPU or PTX of GPU), we prompt the LLMs to use languages they familiar as bridge to write scripts to print kernels.
% The prompt contains the sketch code, the kernel structure, optimization to be implemented and few-shots examples, along with optimization-related hints and target hardware information obtained through Step 1.
% For example, to generate a computation kernel with pipeline optimization for CPU, the structure part describes the vector register definition, the layout of Load, Compute, and Store instruction block.
% Along with the target C910 CPU information (the type and usage of assembly INST, number and width of VR, and so on) in prompt, the LLMs write a Python script to print the assembly-level kernel implementation.
% While for GPU, the LLMs write CuTe code to implement CUDA kernel with PTX-level optimization to fully utilize the peak performance of tensor core.
% To enhance sketch parameter initialization for the next Auto-Tuning module and correctness of assembly kernel, they are joint compiled to test on the target hardware to obtain feedback for refining the sketch and kernel generation scripts.%and adjust parameters using LLMs in a few cycles.

\subsection{Auto-Tuning}
% The kernel and sketch code obtained ensures proper optimization techniques on tensor operator at target hardware.
% Auto-Tuning module is to optimized the parameters of the sketch and the instruction order for further enhancing the performance, as it can uncover optimization opportunities that are difficult to identify during the generation process.
% It integrates a general search algorithm (MCTS) with LLMs to address the lack of semantic-level priors for individual tuning actions, and leaverage search history to guide the exploration for both efficiency and performance.
% The generated sketch and kernel codes guarantee appropriate optimization for tensor operators on the target hardware.
Auto-Tuning aims to optimize the sketch parameters and instruction orders to uncover subtle optimization opportunities that are even hard to be identified by human experts.
It integrates MCTS with LLMs to tackle the shortage of semantic priors for individual tuning actions and leverages search history to guide exploration for efficiency and performance.
% To address the lack of semantic-level priors for individual tuning actions, we developed an auto-tuning module that integrates a general search algorithm (MCTS) with LLMs.
% The LLMs analyze and summarize the search history to guide the exploration of MCTS, enhancing both efficiency and outcomes.

The MCTS nodes represent the current implementation of kernels and sketches.
% The search space involves parameters of sketch and the order of instructions in kernels, so there are two kinds of actions: 1) add or sub the parameters in sketch, like tiling sizes of matrix; 2) rearranging memory access or computation order that are uncorrelated in one pipeline block.
The search space encompasses sketch parameters and the instruction order within kernels.
Consequently, there are two types of actions: 1) adjusting parameters in the sketch, such as matrix tiling sizes, through addition or subtraction; 2) reordering memory access or computation independent within a single instruction block.

As shown in Figure \ref{fig:overview}, the process iterates 4 steps: 1). \textbf{Selection} uses the UCB algorithm to identify a node for expansion; 2). \textbf{Expansion} dynamically grows by utilizing LLMs based on search history ; 3). \textbf{Simulation} tests the tensor operator of the newly expanded node to obtain an initial performance value; 4). \textbf{Backpropagation} updates node values and visit counts along the path. The pseudo-code is provided in the Appendix B.

The LLMs-driven expansion comprises \textbf{path history-based tuning action selection} and \textbf{global history-based tuning space generation}.
% 1). When reaches a node to be expanded, LLMs takes the expansion actions and performance feedback along the search path as context, then are prompted to outputs the most valuable fine-tuning action to expand a new node.
% For example, if the tiling size tuning in the $BN$ dimension significantly impacts performance in the search path, the LLMs tends to select tuning actions in the same dimension.
When reaching a node, take the history tuning actions and performance feedback along the path as context.
LLMs are prompted to output the most valuable tuning action from candidates for expansion.
%For instance, if the size tuning in BN dimension greatly affects performance, LLMs tend to choose tuning actions in the same dimension.
For instance, LLMs tend to select tuning parameters that have a greater impact on performance in the path history, such as the tiling size of \text{BN}.
% 2). After expand a new node, LLMs takes the tuning constraint description, and the performance feedback of all expanded actions in the tree as context, and are prompted to output the legal and potential candidate tuning actions for the new node.
% For example, if most nodes achieve stable performance improvements when the tiling size granularity increases by $32$, the LLMs tends to generate a similar granularity action space, like $\{BN+32, BM+32, BK+32, ...\}$.
After expansion, LLMs take the tuning constraint description and the performance feedback of all expanded actions in the tree as context and output legal and potential candidate tuning actions for the new node.
For example, if most nodes achieve stable performance improvements when the tiling size granularity increases by $32$, LLMs tend to generate a similar granularity action space, $\{\text{BN}+32, \text{BM}+32, \text{BK}+32, ...\}$.
% In summary, fine-graind auto-tuning is challenging because lacking expert priors, but \name enables LLMs using history trajectories to automatically reason and dynamically adjust the MCTS algorithm's exploration direction, so that can effectively improve tuning efficiency and performance.
In summary, fine-tuning is challenging due to the absence of expert priors. However, \name enables LLMs to utilize history trajectories for automatic reasoning and dynamic adjustment of search, thereby effectively enhancing efficiency and performance.

% GEMM主实验表
\begin{table*}[!h]
\centering
\scalebox{0.69}{
\renewcommand{\arraystretch}{1.25} % 设置行间距为原来的1.5倍
\begin{tabular}{c|l|rrrr|rrrr}

\toprule
\multirow{2}{*}{Hardware} & \multirow{2}{*}{Method}  & \multicolumn{4}{c|}{ $(m = k = n)$ }  & \multicolumn{4}{c}{$(m, k, n)$}\\
\cmidrule{3-6}
\cmidrule{7-10}
 & 
% & \begin{tabular}[c]{@{}c@{}}512/\\512/\\512 \\ \end{tabular}  
& \small{512}
% & \begin{tabular}[c]{@{}c@{}}1024/\\1024/\\1024 \\ \end{tabular}  
& \small{1024}
% & \begin{tabular}[c]{@{}c@{}}2048/\\2048/\\2048 \\ \end{tabular}  
& \small{2048}
% & \begin{tabular}[c]{@{}c@{}}4096/\\4096/\\4096 \\ \end{tabular}  
& \multicolumn{1}{r|}{\small{4096}}
% & \begin{tabular}[c]{@{}c@{}}512/\\4096/\\4096 \\ \end{tabular} 
& \small{(512,4096,4096)}
% & \begin{tabular}[c]{@{}c@{}}768/\\4096/\\4096 \\ \end{tabular} 
& \small{(768,4096,4096)}
% & \begin{tabular}[c]{@{}c@{}}1024/\\4096/\\4096 \\ \end{tabular} 
& \small{(1024,4096,4096)}
% & \begin{tabular}[c]{@{}c@{}}2048/\\4096/\\4096 \\ \end{tabular} 
& \small{(2048,4096,4096)}\\ 
\midrule 
\multirow{5}{*}{\begin{tabular}[c]{@{}c@{}}C908\\(RISC-V) \\ 
\end{tabular}}
& GPT-4o & 0.04 & 0.02 & 0.02 & 0.02 & 0.02  & 0.03  & 0.02  & 0.02 \\
&\cellcolor[rgb]{0.925,0.925,0.925} +Ours      
&\cellcolor[rgb]{0.925,0.925,0.925} \textbf{8.70($\uparrow$ 218×)} &\cellcolor[rgb]{0.925,0.925,0.925} \textbf{9.16($\uparrow$ 458×)} &\cellcolor[rgb]{0.925,0.925,0.925} 8.78($\uparrow$ 439×) &\cellcolor[rgb]{0.925,0.925,0.925} \textbf{9.23($\uparrow$ 462×)} &\cellcolor[rgb]{0.925,0.925,0.925} 8.33($\uparrow$ 417×) &\cellcolor[rgb]{0.925,0.925,0.925} 7.83($\uparrow$ 261×) &\cellcolor[rgb]{0.925,0.925,0.925} \textbf{9.40($\uparrow$ 470×)} &\cellcolor[rgb]{0.925,0.925,0.925} \textbf{9.45($\uparrow$ 473×)} \\
\cmidrule{2-10} 
&DS-V3 & 0.03 & 0.01 & 0.01 & 0.01 & 0.03  & 0.03  & 0.03  & 0.02 \\
&\cellcolor[rgb]{0.925,0.925,0.925} +Ours 
&\cellcolor[rgb]{0.925,0.925,0.925} 7.97($\uparrow$ 266×) &\cellcolor[rgb]{0.925,0.925,0.925} 7.5($\uparrow$ 750×) &\cellcolor[rgb]{0.925,0.925,0.925} 8.62($\uparrow$ 862×) &\cellcolor[rgb]{0.925,0.925,0.925} 8.97($\uparrow$ 897×) &\cellcolor[rgb]{0.925,0.925,0.925} \textbf{8.57($\uparrow$ 286×)} &\cellcolor[rgb]{0.925,0.925,0.925} \textbf{9.83($\uparrow$ 328×)} &\cellcolor[rgb]{0.925,0.925,0.925} 8.35($\uparrow$ 278×) &\cellcolor[rgb]{0.925,0.925,0.925} 9.25($\uparrow$ 463×)  \\
\cmidrule{2-10} 
% &  & TVM        & 	&	&	&	&	&	&       \\
& OBLAS   & 7.98 & 8.34 & \textbf{8.85} & 9.00 & 8.28  & 8.70  & 8.51  & 9.00       \\ 

\cmidrule{1-10} 
\multirow{5}{*}{\begin{tabular}[c]{@{}c@{}}C910\\(RISC-V) \\ 
\end{tabular}} 
&GPT-4o & 0.18 & 0.14 & 0.1 & 0.09 & 0.15  & 0.16  & 0.16  & 0.16 \\
&\cellcolor[rgb]{0.925,0.925,0.925} +Ours
&\cellcolor[rgb]{0.925,0.925,0.925} \textbf{11.21($\uparrow$ 62×)} &\cellcolor[rgb]{0.925,0.925,0.925} 11.21($\uparrow$ 80×) &\cellcolor[rgb]{0.925,0.925,0.925} \textbf{10.94($\uparrow$ 109×)} &\cellcolor[rgb]{0.925,0.925,0.925} 9.21($\uparrow$ 102×) &\cellcolor[rgb]{0.925,0.925,0.925} \textbf{11.48($\uparrow$ 77×)} &\cellcolor[rgb]{0.925,0.925,0.925} \textbf{11.74($\uparrow$ 73×)} &\cellcolor[rgb]{0.925,0.925,0.925} 11.56($\uparrow$ 72×) &\cellcolor[rgb]{0.925,0.925,0.925} \textbf{11.05($\uparrow$ 69×)}  \\
\cmidrule{2-10} 
&DS-V3 & 0.09 & 0.05 & 0.03 & 0.03 & 0.81  & 0.82  & 0.82  & 0.82 \\
&\cellcolor[rgb]{0.925,0.925,0.925} +Ours
&\cellcolor[rgb]{0.925,0.925,0.925} 10.58($\uparrow$ 118×) &\cellcolor[rgb]{0.925,0.925,0.925} \textbf{11.48($\uparrow$ 230×)} &\cellcolor[rgb]{0.925,0.925,0.925} 10.84($\uparrow$ 361×) &\cellcolor[rgb]{0.925,0.925,0.925} \textbf{9.47($\uparrow$ 316×)} &\cellcolor[rgb]{0.925,0.925,0.925} 11.36($\uparrow$ 14×) &\cellcolor[rgb]{0.925,0.925,0.925} 11.67($\uparrow$ 14×) &\cellcolor[rgb]{0.925,0.925,0.925} \textbf{11.68($\uparrow$ 14×)} &\cellcolor[rgb]{0.925,0.925,0.925} 10.89($\uparrow$ 13×)  \\
% &  & TVM        &	&	&	&	&	&	&       \\
\cmidrule{2-10} 
& OBLAS   & 5.91 & 5.85 & 4.9 & 4.88 & 4.57  & 4.90  & 5.37  & 5.41       \\ 

\cmidrule{1-10} 
\multirow{5}{*}{\begin{tabular}[c]{@{}c@{}}K1\\(RISC-V) \\ 
\end{tabular}} 
& GPT-4o& 0.32 & 0.28 & 0.28 & 0.19 & 0.50  & 0.52  & 0.48  & 0.47 \\
&\cellcolor[rgb]{0.925,0.925,0.925} +Ours
&\cellcolor[rgb]{0.925,0.925,0.925} 9.97($\uparrow$ 31×) &\cellcolor[rgb]{0.925,0.925,0.925} 9.18($\uparrow$ 33×) &\cellcolor[rgb]{0.925,0.925,0.925} 9.53($\uparrow$ 34×) &\cellcolor[rgb]{0.925,0.925,0.925} 9.9($\uparrow$ 52×) &\cellcolor[rgb]{0.925,0.925,0.925} \textbf{11.43($\uparrow$ 23×)} &\cellcolor[rgb]{0.925,0.925,0.925} 10.74($\uparrow$ 21×) &\cellcolor[rgb]{0.925,0.925,0.925} \textbf{10.78($\uparrow$ 22×)} &\cellcolor[rgb]{0.925,0.925,0.925} \textbf{10.7($\uparrow$ 23×)} \\
\cmidrule{2-10} 
&DS-V3 & 0.36 & 0.33 & 0.31 & 0.23 & 0.45  & 0.47  & 0.46  & 0.45 \\
&\cellcolor[rgb]{0.925,0.925,0.925} +Ours
&\cellcolor[rgb]{0.925,0.925,0.925} \textbf{10.34($\uparrow$ 29×)} &\cellcolor[rgb]{0.925,0.925,0.925} \textbf{9.74($\uparrow$ 30×)} &\cellcolor[rgb]{0.925,0.925,0.925} \textbf{10.29($\uparrow$ 33×)} &\cellcolor[rgb]{0.925,0.925,0.925} \textbf{11.74($\uparrow$ 51×)} &\cellcolor[rgb]{0.925,0.925,0.925} 10.48($\uparrow$ 23×) &\cellcolor[rgb]{0.925,0.925,0.925} \textbf{10.87($\uparrow$ 23×)} &\cellcolor[rgb]{0.925,0.925,0.925} 10.14($\uparrow$ 22×) &\cellcolor[rgb]{0.925,0.925,0.925} 10.41($\uparrow$ 23×)  \\
\cmidrule{2-10} 
% &  & TVM        &	&	&	&	&	&	&       \\
& OBLAS   & 4.12 & 4.19 & 4.46 & 4.76 & 4.57  & 4.90  & 5.37  & 5.41       \\ 

% \toprule
% Platform & \multicolumn{1}{c|}{Hardware} & \multicolumn{1}{c|}{Method} & 256   & 512   & 768   & 1024  & 1536  & 2048   & 4096   \\     
\cmidrule{1-10} 
\multirow{7}{*}{\begin{tabular}[c]{@{}c@{}}A72\\(ARM) \\ 
\end{tabular}}        
&GPT-4o & 0.14 & 0.12 & 0.10 & 0.09 & 0.09  & 0.09  & 0.10  & 0.09 \\
&\cellcolor[rgb]{0.925,0.925,0.925} +Ours
&\cellcolor[rgb]{0.925,0.925,0.925} 11.25($\uparrow$ 80×) &\cellcolor[rgb]{0.925,0.925,0.925} 12.63($\uparrow$ 105×) &\cellcolor[rgb]{0.925,0.925,0.925} 12.76($\uparrow$ 128×) &\cellcolor[rgb]{0.925,0.925,0.925} 12.78($\uparrow$ 142×) &\cellcolor[rgb]{0.925,0.925,0.925} \textbf{12.60($\uparrow$ 140×)} &\cellcolor[rgb]{0.925,0.925,0.925} 12.47($\uparrow$ 139×) &\cellcolor[rgb]{0.925,0.925,0.925} \textbf{12.72($\uparrow$ 127×)} &\cellcolor[rgb]{0.925,0.925,0.925} 12.53($\uparrow$ 139×) \\
\cmidrule{2-10} 
&DS-V3 & 0.12 & 0.09 & 0.02 & 0.01 & 0.13  & 0.13  & 0.12  & 0.12 \\
&\cellcolor[rgb]{0.925,0.925,0.925} +Ours
% & 11.39 & 11.92 & 12.43 & 12.91 & 10.92 & 12.62 & 12.67 & 12.69  \\
&\cellcolor[rgb]{0.925,0.925,0.925} 11.39($\uparrow$ 95×) &\cellcolor[rgb]{0.925,0.925,0.925} 11.92($\uparrow$ 132×) &\cellcolor[rgb]{0.925,0.925,0.925} 12.43($\uparrow$ 622×) &\cellcolor[rgb]{0.925,0.925,0.925} \textbf{12.91($\uparrow$ 1291×)} &\cellcolor[rgb]{0.925,0.925,0.925} 10.92($\uparrow$ 84×) &\cellcolor[rgb]{0.925,0.925,0.925} \textbf{12.62($\uparrow$ 97×)} &\cellcolor[rgb]{0.925,0.925,0.925} 12.67($\uparrow$ 106×) &\cellcolor[rgb]{0.925,0.925,0.925} \textbf{12.69($\uparrow$ 106×)} \\
\cmidrule{2-10} 
% & &TVM          & &  &  &  &  &  &  \\ 
% &TVM    & & & & &  &  &  \\
% \cmidrule{2-9}
&OBLAS    & 11.07 & 12.13 & 12.4 & 12.52 & 12.23  & 12.35  & 12.44  & 12.50\\
\cmidrule{2-10} 
&ACL         & \textbf{12.17} & \textbf{12.81} & \textbf{12.98} & 12.63 & 12.57  & 12.54  & 12.58  & 12.62 \\

\cmidrule{1-10} 
\multirow{8}{*}{\begin{tabular}[c]{@{}c@{}}A76\\(ARM) \\ 
\end{tabular}} 
& GPT-4o & 0.35 & 0.33 & 0.27 & 0.17 & 0.34  & 0.39  & 0.32  & 0.27 \\
&\cellcolor[rgb]{0.925,0.925,0.925} +Ours
&\cellcolor[rgb]{0.925,0.925,0.925} \textbf{34.46($\uparrow$ 98×)} &\cellcolor[rgb]{0.925,0.925,0.925} 34.81($\uparrow$ 105×) &\cellcolor[rgb]{0.925,0.925,0.925} \textbf{36.77($\uparrow$ 136×)} &\cellcolor[rgb]{0.925,0.925,0.925} \textbf{37.31($\uparrow$ 219×)} &\cellcolor[rgb]{0.925,0.925,0.925} \textbf{33.84($\uparrow$ 100×)} &\cellcolor[rgb]{0.925,0.925,0.925} \textbf{33.77($\uparrow$ 87×)} &\cellcolor[rgb]{0.925,0.925,0.925} \textbf{35.72($\uparrow$ 112×)} &\cellcolor[rgb]{0.925,0.925,0.925} 35.32($\uparrow$ 131×)  \\
\cmidrule{2-10}
&DS-V3 & 0.22 & 0.04 & 0.04 & 0.04 & 0.29  & 0.29  & 0.29  & 0.28 \\
&\cellcolor[rgb]{0.925,0.925,0.925} +Ours
&\cellcolor[rgb]{0.925,0.925,0.925} 33.91($\uparrow$ 154×) &\cellcolor[rgb]{0.925,0.925,0.925} \textbf{35.70($\uparrow$ 893×)} &\cellcolor[rgb]{0.925,0.925,0.925} 36.05($\uparrow$ 901×) &\cellcolor[rgb]{0.925,0.925,0.925} 36.67($\uparrow$ 917×) &\cellcolor[rgb]{0.925,0.925,0.925} 32.72($\uparrow$ 113×) &\cellcolor[rgb]{0.925,0.925,0.925} 33.48($\uparrow$ 115×) &\cellcolor[rgb]{0.925,0.925,0.925} 35.66($\uparrow$ 123×) &\cellcolor[rgb]{0.925,0.925,0.925} \textbf{36.43($\uparrow$ 130×)}  \\
\cmidrule{2-10}
% & &TVM          &	&	&	&	&	&	&       \\
&TVM    & 33.79 & 33.57 & 32.99 & Failed & 29.56  & 28.92  & 27.17  & Failed  \\
\cmidrule{2-10}
&OBLAS    & 27.97 & 31.25 & 33.48 & 34.27 & 30.01  & 31.25  & 32.37  & 33.95     \\ 
\cmidrule{2-10}
&ACL         & 34.08 & 32.44 & 32.43 & 30.82 & 32.20  & 31.86  & 31.64  & 31.10    \\

% \bottomrule
\cmidrule{1-10} 

\cmidrule{1-10} 
% \toprule
% \multicolumn{1}{c|}{} & \multicolumn{1}{c|}{Method}   
\multirow{2}{*}{Hardware} & \multirow{2}{*}{Method}  & \multicolumn{4}{c|}{ $(m = k = n)$ }  & \multicolumn{4}{c}{$(m, k, n)$}\\
\cmidrule{3-6}
\cmidrule{7-10}
 & 
& \small{2048}
& \small{4096}
& \small{8192}
& \multicolumn{1}{r|}{\small{16384}}
& \small{(16384,8192,1280)}
& \small{(16384,1024,8192)}
& \small{(16384,8192,7168)}
& \small{(16384,3584,8192)}
% & \begin{tabular}[c]{@{}c@{}}2048/\\2048/\\2048 \\ \end{tabular}   
% & \begin{tabular}[c]{@{}c@{}}4096/\\4096/\\4096 \\ \end{tabular}    
% & \begin{tabular}[c]{@{}c@{}}8192/\\8192/\\8192 \\ \end{tabular}    
% & \begin{tabular}[c]{@{}c@{}}16384/\\16384/\\16384 \\ \end{tabular} 
% & \begin{tabular}[c]{@{}c@{}}16384/\\8192/\\1280 \\ \end{tabular} 
% & \begin{tabular}[c]{@{}c@{}}16384/\\1024/\\8192 \\ \end{tabular} 
% & \begin{tabular}[c]{@{}c@{}}16384/\\8192/\\7168 \\ \end{tabular} 
% & \begin{tabular}[c]{@{}c@{}}16384/\\3584/\\8192 \\ \end{tabular} 
\\
\cmidrule{1-10} 
%  & \multirow{4}{*}{\begin{tabular}[c]{@{}c@{}}RTX4070 \\ \end{tabular}}        
% & Ours(GPT-4o)           & 1.20 & 1.33 & 1.37 & 1.42 &  & 1.35  & 1.34  \\
% & &Ours(DS-V3)       & 1.20 & 1.33 & 1.37 & 1.42 &  & 1.35  & 1.34  \\
% % & &TVM        & &  &  &  &  &  &  \\ 
% &  &cuBLAS & 0.58 & 0.59 & 0.59 & 0.59 &  & 0.58 & 0.58 \\
% \cmidrule{2-10} 
\multirow{6}{*}{\begin{tabular}[c]{@{}c@{}}RTX\\4060 \\(with \\ CUDA \\ core) \end{tabular}} 
& GPT-4o& 1.2 & 1.1 & 1.1 & 1.1 & 1.12  & 1.18  & 1.00  & 1.10  \\
&\cellcolor[rgb]{0.925,0.925,0.925} +Ours
&\cellcolor[rgb]{0.925,0.925,0.925} 7.84($\uparrow$ 7×) &\cellcolor[rgb]{0.925,0.925,0.925} \textbf{8.08($\uparrow$ 7×)} &\cellcolor[rgb]{0.925,0.925,0.925} \textbf{7.84($\uparrow$ 7×)} &\cellcolor[rgb]{0.925,0.925,0.925} \textbf{7.8($\uparrow$ 7×)} &\cellcolor[rgb]{0.925,0.925,0.925} 8.14($\uparrow$ 7×) &\cellcolor[rgb]{0.925,0.925,0.925} 7.63($\uparrow$ 6×) &\cellcolor[rgb]{0.925,0.925,0.925} 7.93($\uparrow$ 8×) &\cellcolor[rgb]{0.925,0.925,0.925} \textbf{8.19($\uparrow$ 7×)}  \\
\cmidrule{2-10} 
&DS-V3 & 1.21 & 1.08 & 1.09 & 1.1 & 1.13  & 1.20  & 1.00  & 1.07 \\
&\cellcolor[rgb]{0.925,0.925,0.925} +Ours
&\cellcolor[rgb]{0.925,0.925,0.925} \textbf{8.1($\uparrow$ 7×)} &\cellcolor[rgb]{0.925,0.925,0.925} 7.99($\uparrow$ 7×) &\cellcolor[rgb]{0.925,0.925,0.925} 7.13($\uparrow$ 7×) &\cellcolor[rgb]{0.925,0.925,0.925} 7.75($\uparrow$ 7×) &\cellcolor[rgb]{0.925,0.925,0.925} \textbf{8.28($\uparrow$ 7×)} &\cellcolor[rgb]{0.925,0.925,0.925} \textbf{7.71($\uparrow$ 6×)} &\cellcolor[rgb]{0.925,0.925,0.925} \textbf{8($\uparrow$ 8×)} &\cellcolor[rgb]{0.925,0.925,0.925} 7.87($\uparrow$ 7×)  \\
\cmidrule{2-10} 
% &  & TVM        & 7.42	&7.07	&6.93	&6.83	&7.09	&6.33	&6.35       \\
&TVM& 7.87 & 6.55 & 6.93 & Failed
& 7.63 & 6.7 & 6.25 & 7.1 \\
\cmidrule{2-10} 
& cuBLAS   & 7.37 & 7.34 & 7.23 & 7.19 & 7.06  & 6.21  & 7.21  & 7.24    \\ 

\cmidrule{1-10} \multirow{6}{*}{\begin{tabular}[c]{@{}c@{}}A100\\ (with \\ tensor \\ core)
\end{tabular}} 
& GPT-4o& 0.81 & 0.86 & 0.88 & 0.74 & 0.75  & 0.85  & 0.80  & 0.74\\
&\cellcolor[rgb]{0.925,0.925,0.925} +Ours
&\cellcolor[rgb]{0.925,0.925,0.925} 183.73($\uparrow$ 227×) &\cellcolor[rgb]{0.925,0.925,0.925} 260.59($\uparrow$ 303×) &\cellcolor[rgb]{0.925,0.925,0.925} 289.13($\uparrow$ 329×) &\cellcolor[rgb]{0.925,0.925,0.925} 293.44($\uparrow$ 397×) &\cellcolor[rgb]{0.925,0.925,0.925} \textbf{278.44($\uparrow$ 371×)} &\cellcolor[rgb]{0.925,0.925,0.925} 237.25($\uparrow$ 279×) &\cellcolor[rgb]{0.925,0.925,0.925} 283.87($\uparrow$ 355×) &\cellcolor[rgb]{0.925,0.925,0.925} 278.04($\uparrow$ 376×) \\ 
\cmidrule{2-10}
&DS-V3  & 14.13 & 17.74 & 17.31 & 18.76 & 18.98  & 18.85  & 18.88  & 18.96  \\
&\cellcolor[rgb]{0.925,0.925,0.925} +Ours
&\cellcolor[rgb]{0.925,0.925,0.925} 183.19($\uparrow$ 13×) &\cellcolor[rgb]{0.925,0.925,0.925} \textbf{262.05($\uparrow$ 15×)} &\cellcolor[rgb]{0.925,0.925,0.925} 290.86($\uparrow$ 17×) &\cellcolor[rgb]{0.925,0.925,0.925} 292.36($\uparrow$ 16×) &\cellcolor[rgb]{0.925,0.925,0.925} 274.88($\uparrow$ 14×) &\cellcolor[rgb]{0.925,0.925,0.925} \textbf{238.26($\uparrow$ 13×)} &\cellcolor[rgb]{0.925,0.925,0.925} 283.66($\uparrow$ 15×) &\cellcolor[rgb]{0.925,0.925,0.925} \textbf{278.48($\uparrow$ 15×)}   \\
\cmidrule{2-10} 
&TVM & 159.66 & 195.59 & 212.79 & 212.48 & 220.33 & 199.8 & 221.02 & 217.76 \\
\cmidrule{2-10} 
& cuBLAS   & \textbf{185.96} & 246.1 & \textbf{292.2} & \textbf{298.44} & 268.58  & 237.19  & \textbf{285.18}  & 253.72      \\ 
\bottomrule
\end{tabular}
}
\caption{GEMM performance comparison on various hardware, LLMs, and matrix dimensions. The first line of each LLM indicates using the vanilla prompt. Performance is measured with GFLOPS and TFLOPS for CPUs and GPUs, respectively. (OBLAS means OpenBLAS, and Failed signifies that the target platform lacks adequate memory for TVM compiling.)}

\label{tab:performance}
\end{table*}
% end GEMM主实验表

\section{Evaluation}

\subsection{Experiment Setup}
% 
% To demonstrate the generality and effectiveness of our method, we select state-of-the-art baselines and four major factors affecting performance for our experiments.

%To validate the performance of the tensor operators generated by \name, as well as the method's generality and efficiency, we selected SOTA libraries and commonly used auto-compilers as baselines and conducted extensive experiments across four key influencing dimensions.
To validate the performance of the tensor operators generated by \name and assess the method's generality and efficiency, we conduct comprehensive evaluations across three different hardware platforms, four distinct LLMs, and two representative tensor operators with varied dimensions. (Due to page limits, more results are shown in Appendix A.)
%(As pages limit, some results refers to Appendix A)

\noindent \textbf{Hardware Platforms.} \name is tested on a wide variety of hardware, including different CPUs with diverse architectures and capabilities (C906, C908, C910\cite{chen2020xuantie} and K1 of RISC-V, A76 and A72 of ARM, and NVIDIA GPU RTX4060 with CUDA Core and A100 with Tensor Core).

\noindent \textbf{LLMs.} The overall performance of \name  is validated with two SOTA LLMs: GPT-4o\cite{gpt4o} (proprietary) and DeepSeek-V3 (DS-V3 for short) \cite{DeepSeekV3} (open-source). 
% Additionally, we conduct more ablation experiments on two  LLMs: 
Two additional LLMs including Claude 3.5 Sonnet-20241022\cite{claude35} and Llama-3.1-405B\cite{lamma31}, are used in ablation experiments.

\noindent \textbf{Benchmarks.} Experiments are conducted on representative tensor operators (GEMM and Conv).
%The GEMM dimensions contain both typical regular size and irregular size derived from Llama7b\cite{llama1} and Llama3 70b\cite{llama3}.
Typical regular and irregular GEMM dimensions  in Llama7b\cite{llama1} and Llama3 70b\cite{llama3} are included.
Typical Conv dimensions in classical CNNs, such as ResNet-50\cite{resnet50}, VGG-16\cite{vgg16} and U-Net\cite{UNet}, are included. 
%The Conv dimension derived from commonly used CNNs: ResNet-50\cite{resnet50}, VGG-16\cite{vgg16} and U-Net\cite{UNet}. 

\noindent \textbf{Comparison Baselines.} \name is compared with vanilla prompt and CoT prompt~\cite{wei2022chain}. Besides, manually optimized libraries (OpenBLAS\cite{openblas} for RISC-V CPUs, ACL / OpenBLAS for ARM CPUs, cuBLAS\cite{cuBLAS} / cuDNN for NVIDIA GPUs), and commonly used auto-compilers (TVM \cite{tvm} for ARM CPUs and NVIDIA GPUs
\footnote{TVM lacks support for RISC-V CPUs.}) are also compared.

\subsection{Overall Performance}
\begin{figure}[htbp]
    \centering
    \includegraphics[width=1.0\linewidth]{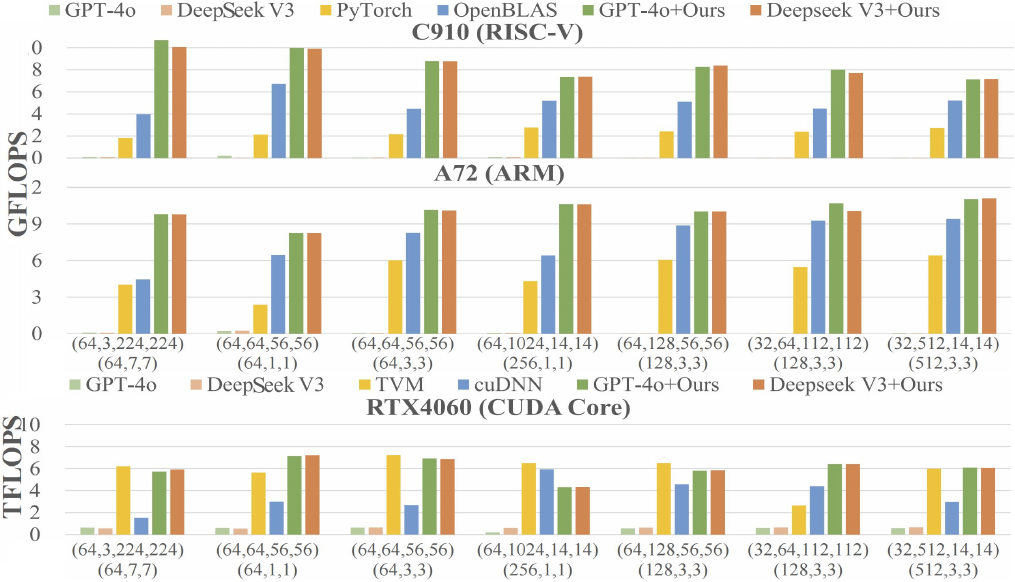}
    \caption{Conv operator performance comparison}% on various hardware, LLMs, and shapes.} %Performance is measured with GFLOPS and TFLOPS for CPU and GPU, respectively. }
    \label{fig:conv_result}
\end{figure}

Table~\ref{tab:performance} and Figure~\ref{fig:conv_result} show the performance of the GEMM and Conv operators generated by \name on various hardware platforms, respectively. The performance is compared with vanilla prompt, manually optimized libraries, and auto-compiler.
% It is clearly evident from the results that \name can enable LLMs to generate high-performance tensor operators across hardware platforms.
The results clearly demonstrate that \name enables LLMs to generate high-performance assembly-level tensor operators across diverse platforms.
% 在不同的硬件平台上，我们的方法生成代码的性能是使用GPT-4o和DeepSeek-V3直接生成代码性能的上百倍。在CPU平台上，我们的方法超过大部分手写库，例如，在RISC-V K1平台，我们的方法是OpenBLAS性能的2.05倍~2.51倍，在ARM A76平台，我们的方法是ACL性能的1.01倍~1.21倍。在GPU平台，我们的方法显著超过TVM，在大部分维度达到cuBLAS性能的90%。

\name outperforms LLMs with vanilla prompt by several orders of magnitude (up to $1291\times$) on GPT-4o and DeepSeek-V3 across various hardware platforms. 
% On RISC-V CPU, our method performs from $60\times$ to $248\times$ better than the vanilla prompt. 
On CPUs, \name surpasses most manually optimized libraries. %up to 251\%. 
% 写整体性能（RISC-V/ARM），不写芯片，单换一行
For instance, on the RISC-V CPUs, \name achieves up to $2.51\times$ performance enhancement over OpenBLAS, while on the ARM CPUs, it achieves at most $1.21$ $\times$ the performance of ACL. 
On GPUs, our method significantly outperforms TVM (up to $1.38\times$ for GEMM and $2.43\times$ for Conv), and achieves comparable or higher performance of cuBLAS and cuDNN in most dimensions (up to $1.24\times$ for GEMM and $3.89\times$ for Conv).
% 加一段话描述即将讲述的三点
We will present the observations of the experiments in the following aspects: LLMs, hardware platforms, and operator dimensions.

% \begin{figure}[htbp]
%     \centering
%     \includegraphics[width=0.9\linewidth]{conv_result.pdf}
%     \caption{Conv operator performance comparison on various hardware, LLMs, and shapes.} %Performance is measured with GFLOPS and TFLOPS for CPU and GPU, respectively. }
%     \label{fig:conv_result}
% \end{figure}

%我们的方法对于大模型本身的依赖
% \textbf{The Impact of LLMs: The performance of tensor operators generated by various LLMs shows substantial discrepancies, whereas our approach attains consistently high performance.}
\textbf{The Impact of LLMs:  \name consistently yields high performance, whereas the vanilla prompt performs poorly.}
%结论：直接用大模型生成的性能差异大，然而使用我们的方法后，都能达到较好的性能
% 我们的实验结果表明，大模型有能力进行原语级别算子生成，但是需要合适的方式激活他。
% 使用不同大模型利用简单提示词生成的代码结构不同，因此性能有所不同。由于不同大模型对于优化技术的理解能力不同，因此生成的代码中优化程度也不相同。例如：在C910芯片上，GPT-4o生成的代码中包含分块这一优化，而DS-V3生成的代码不包含，因此它们的性能相差三倍。
% Due to the varying abilities of LLMs, the generated tensor operator differs in optimization levels and exhibit significant performance variations. 
% For example, on the C910 chip, the tensor operator generated by GPT-4o incorporates tiling optimization, whereas DeepSeek-V3 does not, yielding a three-fold performance disparity. 
% Our method enables different LLMs to produce consistent optimizations on hardware, with minimal demands on the models' inherent capabilities, showcasing its robustness. 
% For example, on C910, the performance between GPT-4o and DeepSeek-V3 is consistent (disparity within 10\%).
\name activates LLMs to automatically generate sketch and hardware-primitive-level kernels with high performance across LLMs.
For example, on C910 CPU, the performance disparity of \name using GPT-4o and DeepSeek-V3 is within 10\%.
% In contrast, the vanilla prompt can't prompt LLMs to generate usable assembly code. Without it, the target hardware's features can't be fully utilized, leading to much lower performance than our method. 
% Moreover, owing to the diverse capabilities of LLMs, the generated tensor operators vary in their optimization levels and display substantial performance variations. 
In contrast, the vanilla prompt can only enable LLMs to generate tensor operators with high-level language (C code), but cannot manipulate hardware resources, thus yielding much lower performance.
Moreover, LLMs of vanilla prompt have obvious performance disparities.
For instance, on C910 CPU, the tensor operator generated by GPT-4o incorporates tiling optimization, while DeepSeek-V3 does not, resulting in a $3\times$ performance disparity.

\textbf{The Impact of Hardware Platforms: \name achieves a more significant performance boost on RISC-V CPUs than on ARM CPUs and GPUs.}
%The RISC-V architecture exhibits high levels of openness and flexibility, presenting challenges for manually optimized libraries in providing specialized optimization, while the official version of TVM even lacks native support for RISC-V CPUs.
As an open-source instruction set architecture, RISC-V CPUs feature a highly flexible and diverse set of instructions and architectures. Thus, it poses significant challenges for manually optimized libraries. Besides, the official version of TVM even lacks native support for RISC-V CPUs.
For instance, on C908, C910, and K1 CPUs, we obtain performance enhancements of $1.10\times$, $2.32\times$, and $2.51\times$ respectively in GEMM operation compared with OpenBLAS.
As commercial processors, ARM and Nvidia hardware vendors have employed human experts to manually optimize the BLAS libraries.
%the architectures of ARM CPUs and NVIDIA GPUs are fixed and unified, more easily allowing manual optimization. 
Our approach still yields competitive performance compared ACL on ARM CPUs ($1.02\times$ to $1.21\times$), and cuBLAS on NVIDIA GPUs ($0.98\times$ to $1.24\times$).
Compared with manually optimized libraries, our approach can rapidly adapt to diverse architectures, achieving high performance across all platforms.

%\textbf{The Impact of Operator Shapes: The larger and more irregular the scale of the matrix, the more prominently \name outperforms auto-compilers and manually optimized libraries.}
\textbf{The Impact of Operator Dimensions: \name outperforms auto-compilers and manually optimized libraries, especially on GEMM with larger or irregular dimensions.}
% 相对于小规模矩阵，在大规模和不规则矩阵上，我们的方法相对于其他方法的提升更大。在ARM A76平台上，当矩阵规模是512×512×512时，我们的方法分别是TVM和ACL的1.02倍和1.01倍，在矩阵规模是1024×4096×4096时，我们的方法分别是TVM,和ACL的1.31倍和1.29倍
% Performance improvements are particularly pronounced for large and irregular matrices. 
% Compared to small-scale matrices, our method achieves greater performance improvements on large-scale and irregular matrices. 
% On the ARM A76 CPU, when the matrix size is 512×512×512, our method is 1.02$\times$ and 1.01$\times$ faster than TVM and ACL, respectively. However, when the matrix size increases to 1024×4096×4096, our method is 1.31 $\times$ and 1.29 $\times$ faster than TVM and ACL, respectively.
%Optimizing large and irregular matrices is difficult because their optimization spaces are vast and complex, making manual tuning (ACL) and automated search (TVM) inefficient.
GEMM of larger or irregular dimensions would pose challenges of elaborately manipulating computation and memory resources, otherwise it may lead to inefficient memory access and lower performance, as we see in the manually optimized library of ACL and auto-compiler of TVM.
In contrast, \name performs more precise and efficient tensor operator generation and 
 optimization at the hardware-primitive-level to fully leverage hardware characteristics, thereby achieving higher performance.
%the power of LLMs to efficiently generate better sketch and hardware-primitive-level kernel, and explore the parameter space by utilizing historical search records.
For instance, when the matrix dimension is $1024\times4096\times4096$ on A76 CPU, our method performance is 1.31 $\times$ and 1.29 $\times$ better than TVM and ACL, respectively. 
% For small matrices, manually optimized libraries and auto-compilers often implement specialized optimizations, resulting in superior performance. 
% For example, the performance of our method achieves $1.02\times$ and $1.01\times$ the baseline performance with TVM and ACL when the matrix shape is $512\times512\times512$ on A76 CPU.
% In Table~\ref{tab:performance}, similar conclusions can also be easily observed on other CPUs and GPUs.

% 大的和不规则的提升更高，
% 大的矩阵和不规则矩阵使用手工优化和自动搜索时优化参数空间大，优化难度高，而我们的方法可以通过LLM实现更快的搜索。在小矩阵上，cuda等手工库会进行特殊优化，因此性能更高。

\subsection{Ablation Study}
% 为了更加深入研究我们方法中各组件的有效性以及方法的效率，考虑到实验成本，我们在一些硬件上进行了 大模型能力、框架组件的有效性、搜索技术、开发代价 的消融实验。
%To  investigate the effectiveness of each component in our method and its efficiency, we conducted following ablation studies:
To analyze the effectiveness of each component in \name and its efficiency, we conduct the following ablation studies.

\textbf{Ablation of Prompts: With just a one-line user prompt, \name demonstrates exceptional performance, while the enhancement from CoT is slight.}
As shown in Table~\ref{tab:cot}, our approach ranges from $26.49\times$ to $ 47.00\times$ compared with vanilla prompt, while the CoT prompt ranges from $1.29\times$ to $2.09\times$.
The results indicate that \name is applicable to different LLMs (even to slightly weaker LLMs) and effectively activates LLMs to generate high-performance tensor operator, showing exceptional robustness and scalability.% of our 

% 在RISC-V的K1平台上，我们在更多能力相对没有那么强的大模型上，进行了naive和cot提示词的实验对比，结果表明我们的方法可以适用于不同的LLM，有效的提升了性能，表三中我们的提升从$26.49\times$~47\times。这个结果充分显示我们方法对于LLM的鲁棒性。
\begin{table}[htbp]
\centering
\scalebox{0.7}{
\renewcommand{\arraystretch}{1.25} % 设置行间距为原来的1.5倍
\begin{tabular}{l|rrr}
\toprule
            & \multicolumn{1}{c}{512} & \multicolumn{1}{c}{1024} & \multicolumn{1}{c}{2048} \\ 
\midrule
GPT-4o  & 0.32   & 0.28      & 0.28 \\
+CoT  & 0.67($\uparrow$ 2.09×) & 0.45($\uparrow$ 1.61×)    & 0.44($\uparrow$ 1.57×)           \\
 %& ($\uparrow$ 2.09×)& ($\uparrow$ 1.61×)&($\uparrow$ 1.57×) \\
\rowcolor[rgb]{0.925,0.925,0.925} +\name & \textbf{9.97($\uparrow$31.16×)}  & \textbf{9.18($\uparrow${32.79×})} &     \textbf{9.53($\uparrow${34.04×})}      \\ 
%$\rowcolor[rgb]{0.925,0.925,0.925} & ($\uparrow$ \textbf{31.16×}) &($\uparrow$ \textbf{32.79×})& \textbf{($\uparrow$ \textbf{34.04×})} \\ 
% Claude 3.5 Sonnet(new)
\midrule
\begin{tabular}[l]{@{}l@{}}Claude 3.5 Sonnet\end{tabular}
% Claude 3.5\footnote{Claude 3.5 Sonnet(20241022)}
  & 0.37  & 0.25   & 0.24          \\
+CoT  & 0.49($\uparrow$ 1.32×)    & 0.38($\uparrow$ 1.52×)      & 0.32($\uparrow$ 1.33×)    \\
\rowcolor[rgb]{0.925,0.925,0.925} +\name              & \textbf{9.80 }($\uparrow$ \textbf{26.49×})         & \textbf{7.87 }($\uparrow$ \textbf{31.48×})    & \textbf{8.95 }($\uparrow$ \textbf{37.29×})       \\  
\midrule
DeepSeek-V3  & 0.36    & 0.33     & 0.31   \\
+CoT   & 0.56($\uparrow$ 1.56×)    & 0.53($\uparrow$ 1.61×)        & 0.40 ($\uparrow$ 1.29×)      \\
\rowcolor[rgb]{0.925,0.925,0.925} +\name              & \textbf{10.34 } ($\uparrow$ \textbf{28.72×})        & \textbf{9.74 }($\uparrow$ \textbf{29.52×})    & \textbf{10.29 } ($\uparrow$ \textbf{33.19×})      \\ 

\midrule
Llama-3.1-405B  &  0.31   &   0.26   &  0.23  \\
+CoT   &  0.49($\uparrow$ 1.58×)   &   0.35($\uparrow$ 1.35×)   &    0.33 ($\uparrow$ 1.43×)   \\
% & ($\uparrow$ 1.58×) & ($\uparrow$ 1.35×) & ($\uparrow$ 1.43×) \\
\rowcolor[rgb]{0.925,0.925,0.925} +\name              & \textbf{9.72 } ($\uparrow$ \textbf{31.35×})         & \textbf{9.57 }   ($\uparrow$ \textbf{36.81×})  & \textbf{10.81 } ($\uparrow$ \textbf{47.00×})      \\ 
\bottomrule
\end{tabular}
}

% \caption{Ablation study results of the prompts comparison between vanilla prompt, CoT, and LLM-HPT on the K1 CPU.}
\caption{GEMM performance (GFLOPS) comparison of vanilla prompt, CoT, and \name on K1 CPU.}
\label{tab:cot}
\end{table}

\begin{figure}[htbp]
    \centering
    \includegraphics[width=1.0\linewidth]{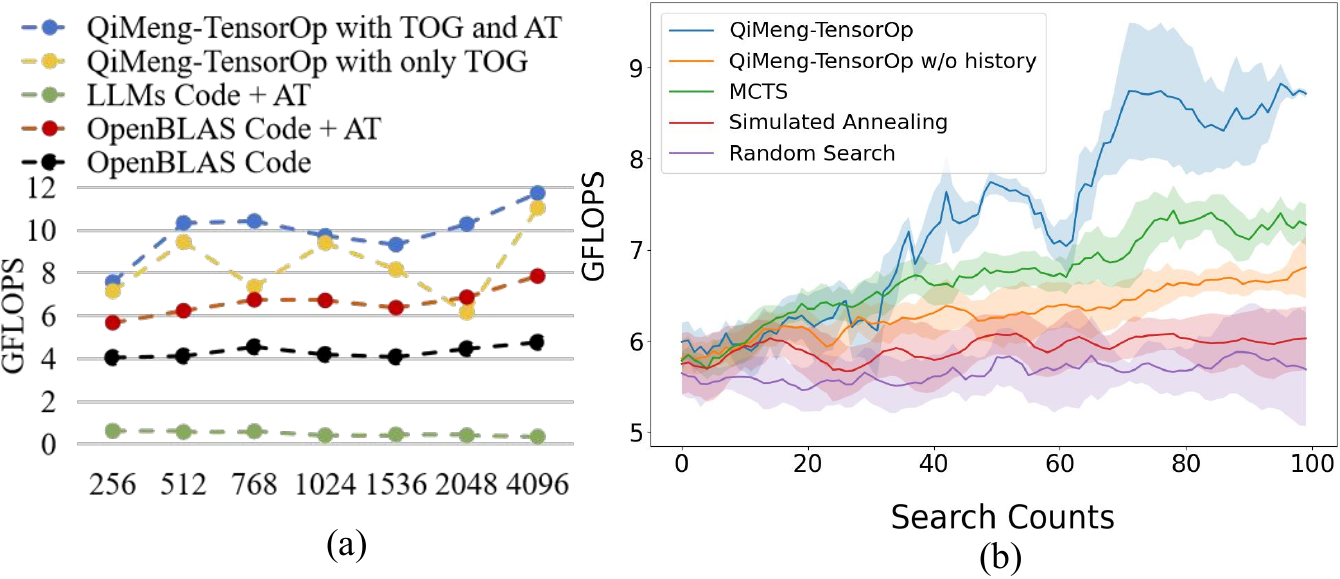}
    \caption{(a) Ablation results of \name components for GEMM on K1 CPU. (b) Performance comparison of tuning methods on GEMM of size 1024 on K1 CPU.}
    % \caption{(a) Performance comparison of tuning methods including \name, \name w/o history, MCTS, Simulated Annealing, and Random Search on GEMM with regular size 1024 (b) Ablation results of \name modules. Both experiments are conducted on K1 CPU.}
    \label{fig:ablation_1}
\end{figure}

\textbf{Ablation of \name Components:}
Figure~\ref{fig:ablation_1}(a) compares the GEMM performance on K1 CPU among five conditions, including \name with both Tensor Operator Generation (TOG) and Auto-Tuning (AT), \name with only TOG, a GPT-4o generated C code with AT, and OpenBLAS with and without AT.
\textbf{(1) Tensor Operator Generation effectively raises performance upper bound.} %xxx数据分析。
%如图4.b中所示，LLM-HPT超过OpenBLAS C Kernel+Auto-Tuning，这也就意味着tensor operator generation生成了汇编层operator能够更好地利用硬件特性，取得超过手工优化的OpenBLAS C Kernel代码。同时，同样是一行prompt，LLM-HPT可以生成高效的汇编代码，vanilla LLMs甚至无法直接生成超过OpenBLAS C Kernel代码的Kernel代码。上述现象说明，tensor operator generation可以更好地激活LLMs对硬件资源的利用，提升生成tensor operator的性能上限。
%\name (blue curve) obviously exceeds LLM C Kernel + AT (green curve) and OpenBLAS C Kernel + AT (red curve), which indicates that the assembly-level kernels generate by TOG can make much better use of hardware characteristics compared with C kernels.
\name (blue curve) obviously exceeds LLM Code + AT (green curve) and OpenBLAS Code + AT (red curve).
As LLMs can only generate C code, and so does the general implementation in OpenBLAS, this indicates that the assembly kernels generated by TOG can better exploit the computing capability of hardware compared with C code.
%The performance TOG comprehensively surpasses OpenBLAS kernel also demonstrate such conclusion.
% thus outperforming the h OpenBLAS C Kernel code.
% Moreover, when given the same one-line user prompt, \name can generate efficient assembly code, while LLMs C Kernel+Auto-Tuning generated by vanilla LLMs performs far worse than the OpenBLAS C Kernel code.
% These phenomena suggest that the tensor operator generation method in \name can help LLMs better utilize hardware resources at hardware-primitive-level, thereby improving the performance ceiling.
%\textbf{(2) Auto-Tuning is vital for ensuring that the tensor operates at its highest level of performance.}
\textbf{(2) Auto-Tuning is vital for superior performance of tensor operates.}
% In the \name w/o Auto-Tuning Module experiment, we directly test the code performance generated by tensor operator generation. 
% Its performance (green curve) fluctuates widely and is inferior to \name evidently. % but still achieves performance comparable to \name in certain cases(256, 1024).
%尽管tensor operator generation生成的汇编kernel的性能远好于openblas c kernel, 但是通过比较LLM-HPT和tensor operator generation可知，在某些矩阵维度上auto-tuning可以发现一些subtle优化机会，进一步增强tensor operator性能。
%Methods with Auto-Tuning, \name (blue curve) and OpenBLAS Code + AT (red curve), significantly outperform the \name with only TOG (yellow curve) and the OpenBLAS code (black curve), respectively.
Auto-Tuning improves the performance of \name with only TOG (yellow curve), as well as OpenBLAS code (black curve), respectively.
This indicates that Auto-Tuning can effectively identify subtle optimization opportunities to achieve superior performance.
Moreover, it has a certain degree of generalization ability for tensor operators generated by different methods. 

\begin{figure}[htbp]
    \centering
    \includegraphics[width=1.0\linewidth]{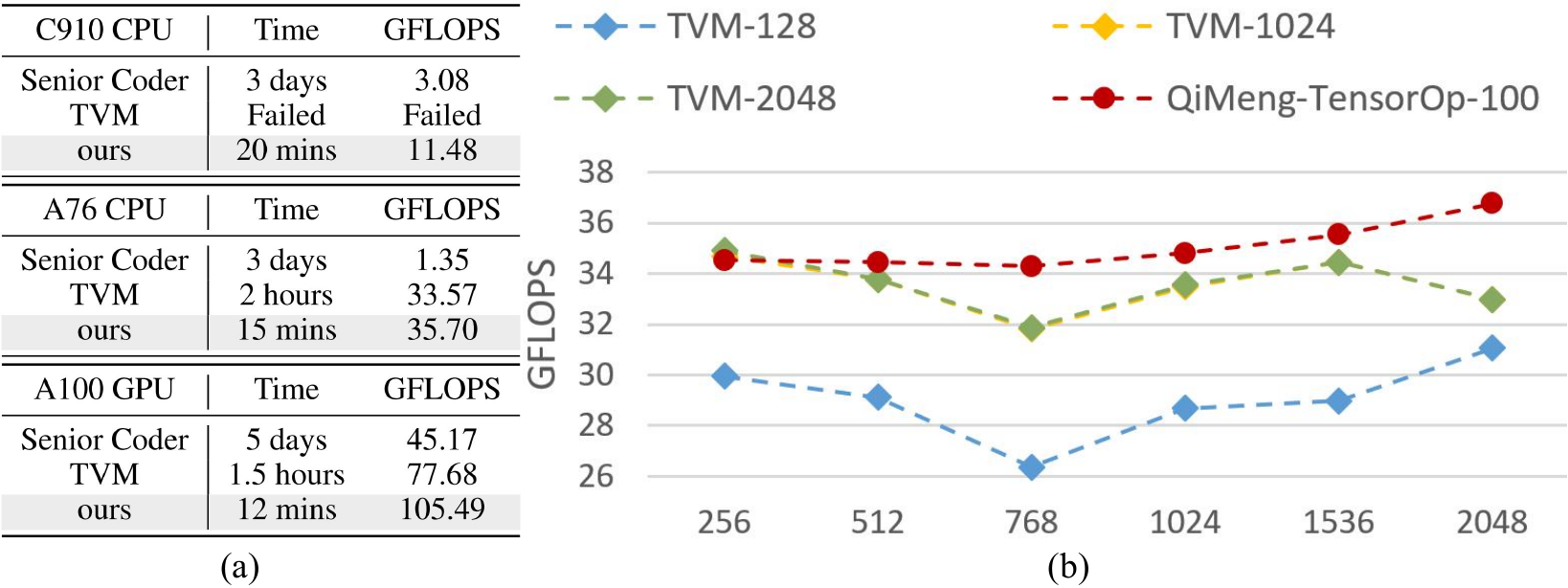}
    \caption{(a) Development cost comparison on GEMM of size 1024 (b) GEMM performance comparison between \name and TVM with different search counts on A76 CPU.}
    % \caption{(a) Comparison of GEMM development cost (b) Search count and performance comparison between \name and auto-compiler on A76 CPU}
    \label{fig:ablation_efficiency}
\end{figure}

\textbf{Ablation of Search Methods: The historical node expansion records is crucial for Auto-Tuning in \name. }
% 历史搜索记录对于Auto-Tuning module in LLM-HPT 非常重要
% LLM-HPT (blue curve) significantly outperforms naive MCTS (green curve) but underperforms naive MCTS without history records (orange curve). In our experiments, we observe that removing history information makes LLM-HPT more prone to suboptimal solutions. Using history can help LLM-HPT avoid repeated errors and optimize its decision strategy.
As shown in Figure~\ref{fig:ablation_1}(b), when search histories are available, \name (blue curve) can leverage them to do dynamic expansion and search, thus exceeding naive MCTS (green curve) in both efficiency and final performance. 
In contrast, \name w/o history (orange curve) degrades to worse than MCTS with fixed expansion.
\textbf{Ablation of Development Cost: (1) \name reduces manually development costs.} 
% Two software engineers with five years of experience serve as senior coders. 
On the three platforms shown in Figure~\ref{fig:ablation_efficiency}(a), our method reduces development costs from several days to no more than twenty minutes compared to software engineers with 5 years of experience by up to $200\times$ (on A100 GPU, 12 mins vs. $5\times8$ hours), while performance is improved by up to 26$\times$ (on A76 CPU). 
\textbf{(2) \name achieves higher performance than TVM with fewer search counts.} 
As shown in Figure~\ref{fig:ablation_efficiency}(b), increasing the number of TVM searches improves code performance, but it has an upper limit.
%In most cases, it lower than the performance of \name with 100 searches.

% 个人经验：TVM对于新硬件很难适应，举个例子：

% 附录：
% 完整的例子
% 多余的结果
% 其他中间结果
\vspace{-10 pt}
\section{Conclusion}
% In this paper, we propose a LLMs framework, \name, automatically generate tensor operators at the hardware primitive level across various platforms, requiring only a single sentence from users to describe the target operator and hardware.
% \name leaverges general hardware intrinsic optimization prior hints and workflow to help LLMs understand hardware and optimization techniques, allowing to automatically extract information from manuals to generate tensor operators with hardware primitives. 147
% Additionally, \name contains a LLM-assisted MCTS algorithm that effectively enhances the efficiency and performance of tuning primitive-level tensor operators on specific hardware
% Extensive evaluations across diverse hardware platforms and tensor operators (GEMM and Conv) of various shapes demonstrate significant performance promotion and development cost reduction.
% In the future, we plan to extend \name to more hardware platforms like deep learning accelerator Cambricon and AMD GPUs, and larger scale of optimization techniques.
In this paper, we present  \name to automatically generate high-performance tensor operators at the hardware primitive level across various platforms, requiring only one user-provided sentence to describe the target operator and hardware.
\name leverages hardware hints and workflow to aid LLMs' comprehension of hardware and optimization, enabling automatic information extraction from manuals for tensor operator generation.
It incorporates an LLM-assisted MCTS algorithm, enhancing the tuning efficiency and performance.
Extensive evaluations on diverse platforms and tensor operators show significant performance and development cost benefits. \name has the potential for continuously evolving hardware architectures.
In the future, we will extend the proposed framework to more operators .
\bibliographystyle{named}
\bibliography{ijcai25}

\end{document}